\documentclass[preprint,12pt]{elsarticle}

\usepackage{color}
\usepackage{float}
\usepackage{multirow}
\usepackage{amsmath}
\usepackage{amssymb}
\usepackage{graphicx}
\usepackage{ulem}
\usepackage{microtype}

\makeatletter

\usepackage{url}
\usepackage{booktabs}
\usepackage{amsfonts}
\usepackage{times}
\usepackage{epsfig}
\usepackage[linesnumbered,ruled,vlined]{algorithm2e}
\usepackage{bookmark}

\@ifundefined{showcaptionsetup}{}{%
 \PassOptionsToPackage{caption=false}{subfig}}
\usepackage{subfig}

\begin{document}

\begin{frontmatter}



\title{Towards Using Count-level Weak Supervision for Crowd Counting}


\author[scu]{Yinjie~Lei}

\ead{yinjie@scu.edu.cn}

\author[scu]{Yan~Liu}

\ead{yanliu27@stu.scu.edu.cn}

\author[dut]{Pingping~Zhang}

\ead{jssxzhpp@mail.dlut.edu.cn}

\author[ua]{Lingqiao~Liu\corref{cor}}

\ead{lingqiao.liu@adelaide.edu.au}

\address[scu]{College of Electronics and Information Engineering, Sichuan University,
Chengdu, China}

\address[dut]{School of Information and Communication Engineering, Dalian University
of Technology, Dalian, China}

\address[ua]{School of Computer Science, The University of Adelaide, Adelaide,
Australia}

\cortext[cor]{Corresponding author}

\begin{abstract}
Most existing crowd counting methods require object location-level
annotation, i.e., placing a dot at the center of an object. While
being simpler than the bounding-box or pixel-level annotation, obtaining
this annotation is still labor-intensive and time-consuming especially
for images with highly crowded scenes. On the other hand, weaker annotations
that only know the total count of objects can be almost effortless
in many practical scenarios. Thus, it is desirable to develop a learning
method that can effectively train models from count-level annotations.
To this end, this paper studies the problem of weakly-supervised crowd
counting which learns a model from only a small amount of location-level
annotations (fully-supervised) but a large amount of count-level annotations
(weakly-supervised). To perform effective training in this scenario,
we observe that the direct solution of regressing the integral of
density map to the object count is not sufficient and it is beneficial
to introduce stronger regularizations on the predicted density map
of weakly-annotated images. We devise a simple-yet-effective training
strategy, namely Multiple Auxiliary Tasks Training (MATT), to construct
regularizes for restricting the freedom of the generated density maps.
Through extensive experiments on existing datasets and a newly proposed
dataset, we validate the effectiveness of the proposed weakly-supervised
method and demonstrate its superior performance over existing solutions.
\end{abstract}

\end{frontmatter}{}


\section{Introduction\label{sec:introduciton}}

Crowd counting aims to estimate the number of object-of-interest within
an input image \cite{zhang2015cross,zhang2016single} or video sequence
\cite{he2019dynamic,Zou2019Enhanced}. The modern solution to the
crowd counting problem is based on the idea of ``learning to count
'' \cite{lempitsky2010learning}, that is, building a learnable
model to directly or indirectly estimate the count number \cite{guerrero2015extremely,xie2018microscopy}.
To train such a crowd counting model, annotation on the object location
is often needed. In the most commonly used annotation protocol, a
dot is put on the object center to create a ``dot map'' to indicate
the object location. Then after convolving with a Gaussian kernel,
a ``dot-map'' can covert to a ``density map'' for model training.
Comparing with the bounding box annotation in object detection \cite{Konyushkova_2018},
or the pixel-level annotation in object segmentation \cite{Pinheiro_2015},
the dot-map based location-level annotation seems to be easier to
obtain. However, for a crowd scene, the number of objects can be large
and in such a scenario collecting the location-level object annotations
can become labor-intensive and time-consuming.

On the other hand, the count-level annotation, i.e., the total count
of objects, can usually be effortlessly obtained in many practical
scenarios. For example, we can take images of the same set of objects
with different spatial arrangements or from different viewpoints and
the total object count remains the same. Once we know the total count
of one image, i.e., ``seed image''\footnote{The total count of seed image can be obtained by using manually counting.
Note that it is common practice to put a dot mark on already counted
objects for the convenience of counting. So the location-level annotation
for the seed images will be naturally obtained as the byproduct of
the counting process.}, the total count for the remaining ones are known\footnote{Note that in this example, we only assume that the camera is movable
at the training stage. The learned crowd counter can be applied to
the scenario that the camera is fixed. Therefore, the multi-view information
or motion information for crowd counting is not always available at
the test stage.}. For the sake of reducing annotation cost, it is desirable to develop
methods that can fully leverage the count-level annotation to \textit{train}
a crowd counting model.

Despite being cheap to collect, count-level annotation induces weaker
supervision signals than the location-level annotation since
the latter naturally derives the count-level annotation but not vice
versa. In the literature, training a model by using a large number
of images with \textit{only count-level annotation} has not been systematically
studied, and it is still unclear to what extent using the count-level
annotation can benefit model training.

\begin{figure}[th]
\begin{centering}
\subfloat[Traditional full supervision setting. \label{fig:sub_a}]{\begin{centering}
\includegraphics[width=6.05cm]{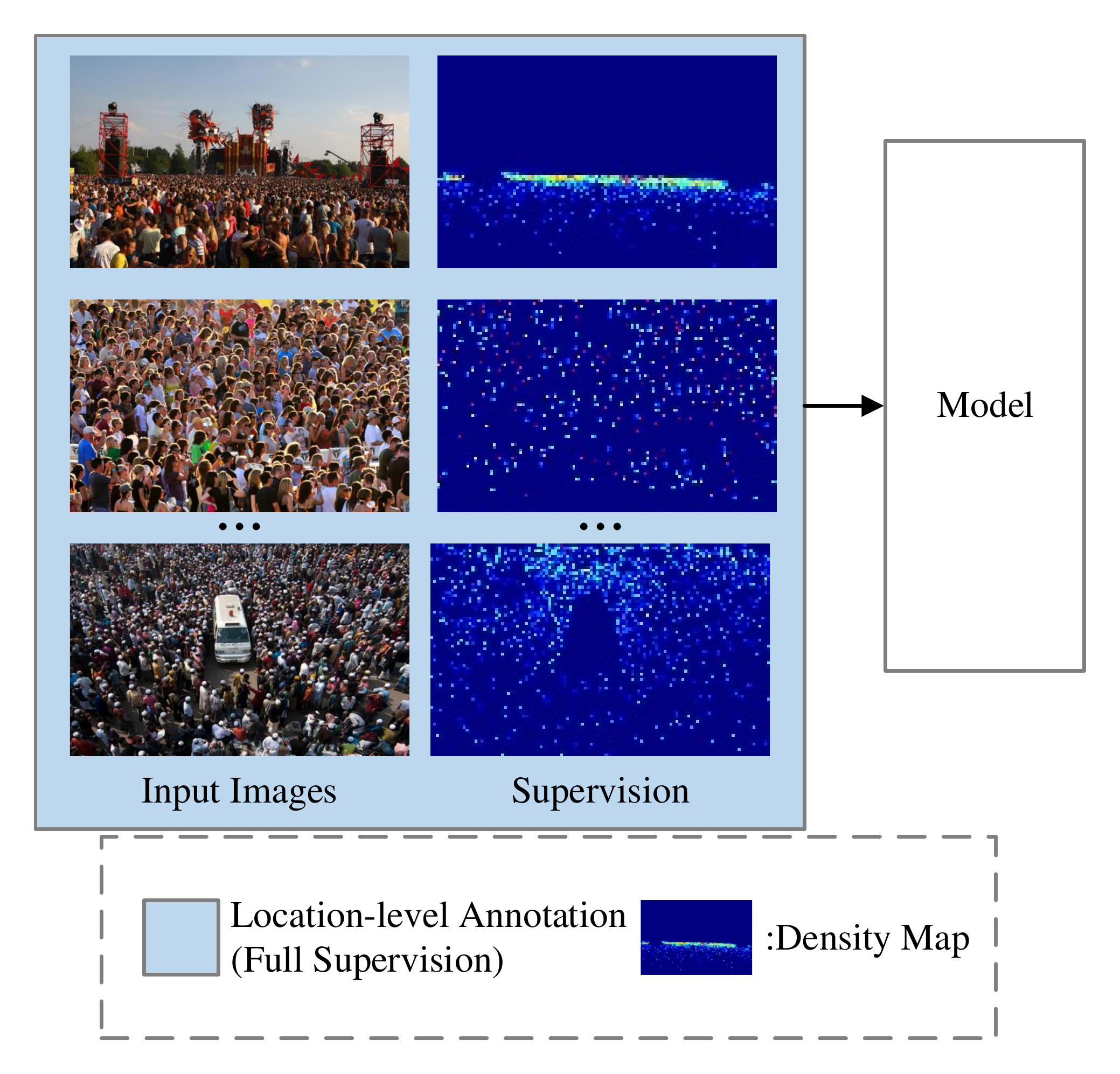}
\par\end{centering}
}\subfloat[Proposed weak supervision setting. \label{fig:sub_b}]{\begin{centering}
\includegraphics[width=6.05cm]{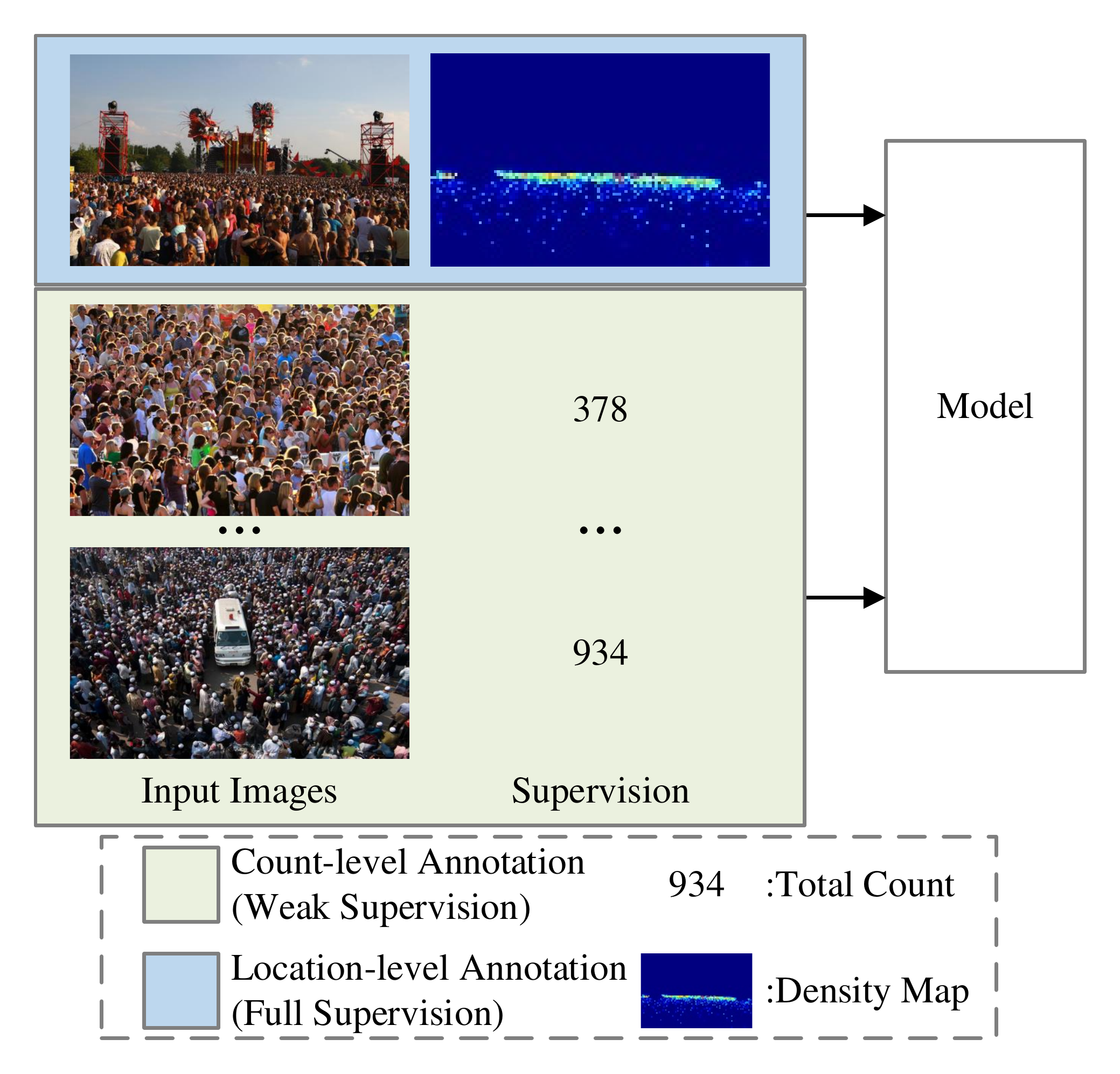}
\par\end{centering}
}
\par\end{centering}
\caption{(a) Traditional full supervision setting. All images are labeled with
location-level annotations. (b) Proposed weak supervision setting.
Only a small number of images are annotated with location-level annotations,
while most images are annotated with the total count. \label{fig:weakly-setting}}
\end{figure}

To fill this gap, this paper studies how to train a crowd counting
model in the weakly-supervised training setting which assumes the
availability of a small amount of location-level annotations and a
large amount of count-level annotation, as demonstrated in Figure
\ref{fig:weakly-setting}. At first glance, training under such weakly-supervised
training setting can be easily achieved with a loss function to encourage
the integration of the estimated density map being close to the ground-truth
total count. However, through our study, we discover that this naive
solution is not sufficiently effective since it does not properly
constrain the predicted density maps with only count-level annotation,
a.k.a, weakly-annotated images\textcolor{red}{.} To overcome this
drawback, we introduce more loss terms to regularize the predicted
density maps. These losses are constructed by adding extra branches
and auxiliary training tasks at the training stage. Specifically,
we introduce multiple auxiliary prediction branches to produce different
but equivalent density maps. Two types of losses are imposed at each
branch: (1) the integral of the predicted density map from each branch
should be close to the object count. (2) density map estimations from
different branches should be consistent. Those auxiliary branches
essentially encourage the feature extractor to encode more accurate
location information to support diverse density map realizations.
Consequently, a better feature extractor and the corresponding crowd
counting model can be obtained.

Furthermore, a new dataset is released to verify that if the count
information obtained effortlessly can benefit the model training and
whether the proposed method is more robust than the naive solution.
Such dataset uses the progressive adding/removing strategy to collect
a large amount of images with easily obtained total count annotation.\textcolor{red}{{}
}We evaluate our method on this newly introduced dataset and the traditional
crowd counting datasets. The results demonstrate the advantage of
the count-level weakly-supervised training and the superior performance
over the baseline methods. The main contributions of this paper can
be summarized as follows:
\begin{itemize}
\item We identify the count-level weakly-supervised method as an effective
way to alleviate the burden of annotations.
\item We develop a novel Multiple Auxiliary Tasks Training strategy to achieve
better prediction performance.
\item We introduce a new dataset that is designed for evaluating the weakly-supervised
crowd counting solutions.
\end{itemize}

\section{Related Works}

\subsection{Crowd Counting by Detection}

Early methods rely on extracting low-level or hand-craft features
\cite{chan2008privacy} to obtain global count in a specific scene,
e.g., Histogram Oriented Gradients (HOG) \cite{dalal2005histograms}
and Haar wavelets \cite{viola2004robust}. Then object detection
algorithms apply to crowd counting and approaches based on detecting
individual heads or body parts \cite{subburaman2012counting} are
widely used. Detection-based methods \cite{ma2015small} are qualified
to sparsely populated regions. Nevertheless, when it comes to some
extreme cases, such as low-resolution, heavy occlusion, counting-by-detection
are impotent to give an appropriate prediction.

\subsection{Crowd Counting by Regression}

To handle images with a dense crowd, the feature-regression-based
methods \cite{chen2012feature} are proposed. These approaches learn
a mapping function from the feature domain to corresponding counts
\cite{lempitsky2010learning} or density \cite{zhang2015cross}.
These methods are relying on the feature extraction mechanism, such
as the Gaussian Process \cite{Chan2010Bayesian,von2016gaussian}
and Random Forests regression \cite{fiaschi2012learning,Pham2015count}.
In the early years, regressing count value is widely adopted. To make
use of spatial information, later works convert this task to a density
map regression problem, and integrating a density map can be used
to obtain the final crowd counts.

\subsection{Crowd Counting by CNN}

Benefited from the deep convolutional neural network \cite{krizhevsky2012imagenet},
counting-by-CNN methods have received much attention. CNN-based methods
\cite{walach2016learning,onoro2016towards,Marsden_2017} typically
center around density map prediction and crowd count regression leveraging
a Fully Convolutional Network (FCN) \cite{long2015fully,badrinarayanan2017segnet,peng2017large}.
Most recent crowd counting methods follow the framework developed
in \cite{lempitsky2010learning}, which solves crowd counting problem
by regressing to the object density map. The regression framework
is more robust than other options such as counting by detection \cite{Ryan2009CrowdCU,Dollar_2017},
counting by segmentation \cite{Wang2009Crowd,kang2014fully}, or
directly estimating the count value \cite{Chan2010Bayesian,Bo2015Bayesian}.

Since then, many works have been proposed to modify this framework.
For example, the work in \cite{Sam_2017} proposes a switching network
architecture to better handle the multi-scale issue in crowd counting.
The context information is exploited in \cite{Sindagi_2017} to produce
more accurate density maps. The state-of-the-art network architecture
\cite{simonyan2014very} in image segmentation is also shown to be
useful for crowd counting in \cite{li2018csrnet:}. This work is
then upgraded by incorporating a two-stream network architecture \cite{zhu2019dual}.
Recently, networks with multiple prediction targets, such as predicting
density maps with different resolutions \cite{Ranjan_2018_ECCV}
or predicting density maps with various smoothness levels \cite{Idrees_2018_ECCV}
are shown to be effective. As shown in \cite{Jeong_2018}, creating
an ensemble of predictors can also boost the prediction accuracy.
Besides, the attention mechanism shows the effectiveness of reducing
the estimation error caused by background noise \cite{liu2019adcrowdnet,jiang2019crowd}.
Incorporating the strength of CNN-based and detection-based solutions
to settle the nonuniform distribution of the crowd is also effective
\cite{Liu_2018,he2019dynamic}. The work in \cite{Liu2019crowd}
proposes a deep structured scale integration network to handle scale
variation. Recently, the spatial information is identified to be effective
to solve the density variation problem \cite{xu2019learn}. Leveraging
foreground and background mask information \cite{jiang2019mask}
is able to improve the robustness and effectiveness of crowd counting
. Besides, a novel Bayesian based loss function \cite{ma2019bayesian}
is proposed to enhance the supervision reliability for crowd counting.\textcolor{red}{{}
}In combination with temporal information, Long Short-Term Memory
(LSTM) is widely used in video crowd counting \cite{Zou2019Enhanced,fang2019locality}.

\subsection{Crowd Counting by Low Supervision Methods}

However, the above works mainly focus on the crowd counting with the
fully-supervised setting, i.e., all images are labeled with location-level
annotations. Recently, some low supervision-based methods (i.e., semi,
weakly or even unsupervised) are proposed with the considerations
of reducing the annotation burden. Few works sought other solutions
to train the crowd counting models under semi-supervised settings
\cite{Liu_2019,Liu_2018_CVPR}. They collect abundant unlabeled crowd
images as extra images of a whole location-level annotated crowd counting
dataset. Besides, they construct a rank loss function on these unlabeled
images to achieve a more accurate prediction. Furthermore, the work
in \cite{von2016gaussian} proposes a weakly-supervised solution
based on the Gaussian process for crowd density estimation. In their
work, all samples are annotated with count-level supervision on the
training set. A novel crowd counting solution leveraging sparse features
is proposed in \cite{sam2019almost} to train a crowd counting model
under an almost unsupervised manner. In their work, most parameters
are trained without any labeled data, and only small part parameters
are updated with location-level annotated data. More recently, the
use of synthetic images is explored to reduce the burden of labor-intensive
annotation \cite{wang2019learning}, which shows the feasibility
of transforming synthetic images to authentic ones. Then, the synthetic
images are labeled with location-level annotations automatically.
Moreover, a GAN-based adaptation method to learn from synthetic images
and the corresponding free density maps is proposed in the work \cite{gao2019featureaware}.
Also, the work \cite{gao2019domainadaptive} constructs a domain
transfer based framework transferring synthetic images to realistic
images to train a crowd counter without any manual label. Those pseudo
labels are produced from a Gaussian-prior Reconstruction.

However, the count-level weakly-supervised crowd counting, i.e., learning
a counting model by using count-level supervision, is still not well-explored.
This paper identifies its potential value in reducing the annotation
workload and proposes a new method to improve the existing methods.
At the methodology level, our work is also related to a recently proposed
semi-supervised learning method \cite{clark2018semisupervised} which
is primarily developed for Natural Language Processing (NLP) applications.
Inspired by the nature of crowd counting, our construction of auxiliary
tasks is significantly diverse from that in \cite{clark2018semisupervised}.

\section{Methodology}

In this section, we first introduce the the traditional location-level
fully-supervised crowd counting setting in Subsection \ref{sec:fully_supervised_crowd_counting}.
Then, Subsection \ref{sec:count_level_weak_supervision_and_problem_formulation}
describes the acquisition of the count-level annotations and the weakly-supervised
crowd counting setting addressed in this paper. In Subsection \ref{sect:naive_solution_for_weakly_supervised_solution}
we discuss a naive solution to weakly-supervised crowd counting. Later,
Subsection \ref{sec:multiple_auxiliary_task_training} elaborates
the details of the proposed Multiple Auxiliary Tasks Training method.
Last, Subsection \ref{sec:asymmetry_training} describes the details
of the introduced asymmetry training strategy.

\subsection{Location-level Fully-supervised Crowd Counting\label{sec:fully_supervised_crowd_counting}}

As described above, most recent crowd counting methods \cite{Sindagi_2017,Tian2018PaDnetPC}
require labeling a dot on each object-of-interest to provide the location-level
annotation. With such full supervision, the crowd counting can be
formulated as a density map regression problem. The density map is
usually obtained by convolving the location-level annotation with
a Gaussian kernel, which is expressed as:

\begin{equation}
D(\mathbf{x})=\sum_{\mathbf{x}_{p}}G(\frac{\|\mathbf{x}-\mathbf{x}_{p}\|^{2}}{\sigma^{2}}),\label{eq:density-map}
\end{equation}
where $\mathbf{x}\in\mathbb{R}^{2}$ denotes the coordinate of a pixel
and $\mathbf{x}_{p}$ denotes the $p$-th annotated point. $G(\frac{\|\mathbf{x}-\mathbf{x}_{p}\|^{2}}{\sigma^{2}})$
indicates a Gaussian kernel with $\mathbf{x}_{p}$ as the mean vector
and $\sigma^{2}$ as the empirically chosen variance term. Note that
since $\int G(\frac{\|\mathbf{x}-\mathbf{x}_{p}\|^{2}}{\sigma^{2}})d\mathbf{x}=1$,
the integral of $D(\mathbf{x})$ equals the total object count. Thus
as long as the density map is accurately estimated by the learned
model, the object count can be readily obtained by taking the integral
of the estimated density map \cite{lempitsky2010learning}.

The Mean Square Error (MSE) loss is widely used in training a density
map predictor, which is formulated as: 
\begin{equation}
\mathcal{L}_{MSE}=\sum_{m\in\mathcal{A}}\int(F(\mathbf{x}|I_{m},\lambda)-D_{m}(\mathbf{x}))^{2}d\mathbf{x},\label{eq:MSE-loss}
\end{equation}
where $F(\mathbf{x}|I_{m},\lambda)$ denotes the density value estimation
at point $\mathbf{x}$ for the $m$-th image given by the model $\lambda$,
which is a deep neural network in our case. For the sake of simplicity,
we use $F(\mathbf{x})$ to denote the estimated density map in the
following part.

\subsection{Count-level Weak Supervision and Problem Formulation\label{sec:count_level_weak_supervision_and_problem_formulation}}

As described in Section \ref{sec:introduciton}, collecting location-level
annotations can be labor-intensive especially for highly crowd scenes.
On the contrary, the total object count can be obtained easily in
many practical scenarios. This subsection reviews several methods
to obtain such count-level annotation: (1) Taking images of the same
set of objects from different viewpoints or shuffling the object arrangement
before taking the image again. This method can generate many images
with identical object count. Once the total count of one image, i.e.,
a ``seed image'', is known, the counts for the remaining images
are known. (2) Similar to (1), but each time removing or adding a
small amount (which is easy to count) of objects before re-taking
the image. This approach can effortlessly generate a sequence of images
with diverse object counts. (3) the total number of objects can be
found out via other measurements, e.g., derived from the total weight
of objects or estimated through other sensors at the offline stage.

Although count-level annotation is cheap to obtain, its supervision
signal is weaker in comparison with the traditional location-level
supervision. To achieve appropriate performance, this paper considers
a semi-supervised alike setting.

\textbf{Our setting}: We first assume there is a small set of images
$\mathcal{A}_{F}$ with a location-level annotation, where image $I^{m}$
in $\mathcal{A}_{F}$ is annotated with the ground-truth density map
$D_{m}$. Then we also assume there is a large set of images $\mathcal{A}_{W}$
with a count-level annotation, where image $I^{n}$ in $\mathcal{A}_{W}$
is annotated with the ground-truth object count $c_{n}$. In the scenario
of taking multiple images of the same set of objects, a small number
of location-level annotations can be obtained as a byproduct of counting
the objects in the ``seed image'', which makes this assumption more
realistic in practical applications. For ease of reading, we summarize
a set of important notations used in this paper as Table \ref{Table:notation-variable}.

\begin{table}
\caption{The notation of variables in this paper. \label{Table:notation-variable}}

\centering{}%
\begin{tabular}{cc}
\hline 
Variables & \multicolumn{1}{c}{Description}\tabularnewline
\hline 
$I^{m}\in\mathbb{R}^{2}$ & the image with location-level annotation\tabularnewline
\hline 
$I^{n}\in\mathbb{R}^{2}$ & the image with count-level annotation\tabularnewline
\hline 
$D_{m}\in\mathbb{R}^{2}$ & the ground truth density map for $I^{m}$\tabularnewline
\hline 
$c_{n}\in\mathbb{R}$ & the ground truth total count for $I^{n}$\tabularnewline
\hline 
$F_{0}(\mathbf{x})\in\mathbb{R}^{2}$ & the predicted density map from primary branch\tabularnewline
\hline 
$F_{k}(\mathbf{x})\in\mathbb{R}^{2}$ & the predicted density map from the $k$-th auxiliary branch\tabularnewline
\hline 
$f_{b}$ & the feature extractor backbone\tabularnewline
\hline 
$g_{0}$ & the primary branch\tabularnewline
\hline 
$g_{1},...,g_{k}$ & the auxiliary branch\tabularnewline
\hline 
$\mathcal{L}$ & the loss function of model\tabularnewline
\hline 
\end{tabular}
\end{table}

\subsection{Naive Solution for Weakly-supervised Crowd Counting\label{sect:naive_solution_for_weakly_supervised_solution}}

For model training in the weakly-supervised setting, it is straightforward
to extend the traditional density map estimation framework by imposing
the integral the estimated density map being close to the ground-truth
object count for count-level annotated images. The overall training
objective function for count-level weakly-supervised crowd counting
can be written as:

\begin{align}
 & \mathcal{L}_{Base}=\mathcal{L}_{MSE}+\alpha\mathcal{L}_{count}=\nonumber \\
 & ~~~~~~~\sum_{m\in\mathcal{A}_{F}}\int(F(\mathbf{x})-D_{m}(\mathbf{x}))^{2}d\mathbf{x}+\alpha\sum_{n\in\mathcal{A}_{W}}|\int F(\mathbf{x})d\mathbf{x}-c_{n}|\label{eq:weakly-supervised}
\end{align}

At first glance, this simple solution seems to be sufficient for training
the crowd counting network. However, we find that it leads to unsatisfying
results in practice. Its major weakness is that the count loss, i.e.,
the second term of Eq.\ref{eq:weakly-supervised}, holds a very weak
constraint on the generated density map. The network can easily achieve
low count-loss by producing less desirable density map which does
not encode the accurate object locations and results in poor generalization
performance.

\subsection{Multiple Auxiliary Tasks Training (MATT)\label{sec:multiple_auxiliary_task_training}}

To overcome the limitation of the naive solution described in Subsection
\ref{sect:naive_solution_for_weakly_supervised_solution}, we introduce
more regularization terms to restrict the freedom of the estimated
density maps on the weakly-supervised images. Our idea is to construct
multiple auxiliary branches in addition to the primary branch which
produces the density map used during the test stage. Both the primary
branch and the auxiliary branches generate density maps. Those density
maps are supposed to be diverse but equivalent, that is, all the density
maps could be derived from the same dot-map (i.e., location-level
annotation) but with different smoothness levels.

\begin{figure}
\centering{}\includegraphics[width=12.25cm]{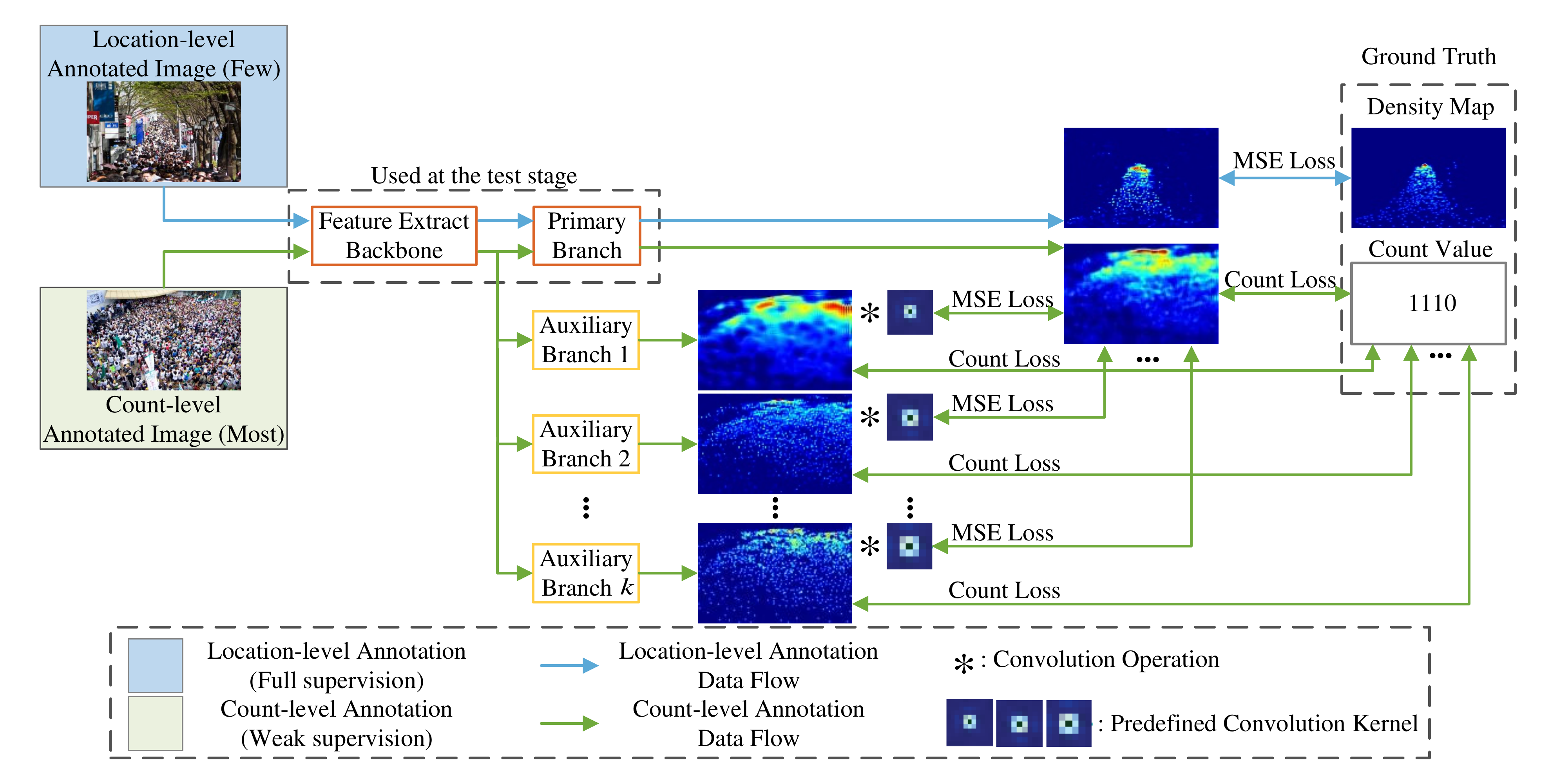}\caption{Overview of the proposed method. For location-level annotated images,
we require the predicted density map close to ground-truth. The proposed
method also generates multiple auxiliary branches. All branches share
a same feature extractor. Once training is done, the auxiliary branches
will be discarded and only the feature extractor and the primary branch
will be used during testing phase. \label{fig:overview-of-the-proposed-method}}
\end{figure}

To this end, we introduce auxiliary tasks in the form of auxiliary
losses to each branch. Specifically, our losses reflect two requirements:
(1) The integral of the predicted density map should be close to the
ground-truth object count. (2) The predicted density maps from an
auxiliary branch should be consistent with the predicted density map
from the primary branch. Rather than directly minimizing the discrepancy
of the predicted density maps between primary and auxiliary branches,
we introduce an additional step to convolve the prediction from each
auxiliary branch with a different predefined Gaussian kernel, and
require convolution result to be close to the density map generated
from the primary branch. The introduction of this extra convolution
operation avoids setting the identical prediction target for each
auxiliary branch. It can prevent the auxiliary branch from collapsing
into the identical regressor as the primary branch. The scheme of
our strategy is illustrated in Figure \ref{fig:overview-of-the-proposed-method}.
Formally, the whole loss function of our method can be written as:

\begin{align}
\mathcal{L}_{aux} & =\beta_{1}\sum_{k}\int\left((F_{k}*h_{k})(\mathbf{x})-F_{0}(\mathbf{x})\right)^{2}d\mathbf{x}+\nonumber \\
 & \beta_{2}\sum_{k}|\int F_{k}(\mathbf{x})d\mathbf{x}-c_{n}|~~~~~~~~k\in\{1,\cdots,K\},\label{eq:auxiliary-loss}
\end{align}
where $K$ is the total number of auxiliary branches. $h_{k}$ is
a predefined convolution kernel for the $k$-th auxiliary branch.
In our implementation, we constructed those convolution kernels by
using a Gaussian function with relatively low variance ($\sigma=1$)
and different kernel sizes, e.g., $3\times3$, $5\times5$, $3\times5$
and $5\times3$ for $h_{1}$ to $h_{4}$, respectively. Note that
the low variance Gaussian function will have significant values outside
the kernel boundaries, so each $h_{k}$ is essentially a truncated
Gaussian kernel. We normalize the values inside a kernel to make their
integral equal to 1. This normalization ensures that the ground-truth
total count of each branch is equal to the total count of the main
branch.\textcolor{red}{{} }$F_{k}$ is the predicted density map from
the $k$-th auxiliary branch $g_{k}$. $F_{0}(\mathbf{x})$ is the
density map estimated from the primary branch $g_{0}$. All the branches
share the same feature extractor $f_{b}$. Our method can be intuitively
understood as follows: a crowd counting network can be decomposed
into a feature extractor and a density map regressor. The feature
extractor produces features that implicitly encode the object location
information and the density map regressor converts those features
into a realization of density map (recall that the density map is
derived from the location annotation and it is not unique). Our method
essentially requires the extracted features to support multiple realizations
of density maps and their relationships. These requirements can enforce
the feature extractor to encode location information more accurately
since by doing so the features can be easily converted to the different
realization of density maps.

\subsection{Asymmetry Training\label{sec:asymmetry_training}}

Directly training the model with the loss in Eq. \ref{eq:auxiliary-loss}
can result in less satisfying results. The reason is that the primary
branch and the auxiliary branches can co-adapt with each other to
produce low consistency loss (i.e., the first term in Eq. \ref{eq:auxiliary-loss}).
To solve this issue, we devise an asymmetry training strategy that
ensures the training signal of the primary branch only comes from
the credible sources, i.e. the ground-truth density map from fully-supervised
data or the ground-truth count from weakly-supervised data. In other
words, when optimizing Eq. \ref{eq:auxiliary-loss}, the parameters
of the primary branch is fixed due to predicted density maps in other
branches are not very reliable. \textit{$F_{0}(\mathbf{x})$ is treated
as a constant in} Eq. \ref{eq:auxiliary-loss}, which means the gradient
will not back-propagate through the primary branch $g_{0}$.

\begin{algorithm}[!t]
\KwIn{Current mini-batch $\mathcal{B}=\{I^{m},D_{m}\}\bigcup\{I^{n},c_{n}\}$, feature extractor backbone $f_{b}$, primary branch $g_{0}$, and auxiliary branches $\{g_{k}\}$. }
\KwOut{Updated $f_{b}$, $g_{0}$ and $\{g_{k}\}$. }
\For{$s$ in $\mathcal{B}$} {\If{$s$ is labeled with the {location-level annotations (density map}) $D_{m}$}{Use $\mathcal{L}_{MSE}=\int(F_{0}(\mathbf{x})-D_{m}(\mathbf{x}))^{2}d\mathbf{x}$ and back-propagate gradients to $f_{b}$ and $g_{0}$} \Else{ {$s$ is} {labeled with count-level annotations (total count) $c_{n}$,} calculate the density map estimation $F_{0}$ from the primary branch \\  Use $\mathcal{L}_{count}=\alpha|\int F_{0}(\mathbf{x})d\mathbf{x}-c_{n}|$ and update $f_{b}$ and $g_{0}$. \\  Frozen the parameters in $F_{0}$ and use $\mathcal{L}_{aux}$ in Eq. \ref{eq:auxiliary-loss} to update $f_{b}$ and $\{g_{k}\}$. } }
\caption{Multiple auxiliary task training.} \label{algo:training}
 \end{algorithm}

This treatment essentially imposes such an \textit{asymmetry learning
strategy}: the auxiliary branches will learn from the primary branch
but not vice versa. This is because we want the auxiliary tasks only
to assist training the feature extractor but not affecting the primary
task training. We postulate that the density map regressor is sensitive
to the input training signal, but the feature extractor is insensitive
to it. Feature extractor training is more robust to the imprecise
training signal since its output can be further adapted by the regressor.
On this basis, the asymmetry training avoids the gradient of the auxiliary
branches from flowing into the primary branch. The overall auxiliary
training objective function for count-level weakly-supervised crowd
counting in our method can be formulated as:

\begin{align}
 & \mathcal{L}_{MATT}=\mathcal{L}_{MSE}+\alpha\mathcal{L}_{count}+\mathcal{L}_{aux}=\nonumber \\
 & ~~~~~~~~~\sum_{m\in\mathcal{A}_{F}}\int(F_{0}(\mathbf{x})-D_{m}(\mathbf{x}))^{2}d\mathbf{x}+\text{}\nonumber \\
 & ~~~~~~~\alpha\sum_{n\in\mathcal{A}_{W}}|\int F_{0}(\mathbf{x})d\mathbf{x}-c_{n}|+\text{}\nonumber \\
 & ~~~~~~~~\beta_{1}\sum_{k}\int\left((F_{k}*h_{k})(\mathbf{x})-F_{0}(\mathbf{x})\right)^{2}d\mathbf{x}+\text{}\nonumber \\
 & ~~~~~~~~\beta_{2}\sum_{k}|\int F_{k}(\mathbf{x})d\mathbf{x}-c_{n}|~~~k\in\{1,\cdots,K\}.\label{eq:auxliary-supervised}
\end{align}

In practice, we use a stochastic gradient descent algorithm to train
the model and the pseudo-code of our methods is shown in Algorithm
\ref{algo:training}. Note that the training procedure for a count-level
annotated sample essentially consists of two steps: the update of
the primary branch and the update of the auxiliary branches.

\section{Multi-Shot Crowd Counting (MSCC) dataset\label{sec:multi_shot_crowd_counting}}

This work advocates the advantage of using count-level annotation
and proposes a new way for weakly-supervised crowd counting. To verify
the effectiveness of the proposed method, we certainly can evaluate
our method on the existing crowd counting datasets and modify their
evaluation protocol to suit the weakly-supervised setting. However,
the total counts of the object in those datasets are not collected
in the way as discussed in Subsection \ref{sec:count_level_weak_supervision_and_problem_formulation}.
One can integral ground-truth density map to obtain total object count
in those datasets. One may wonder whether the samples collected in
a way in Subsection \ref{sec:count_level_weak_supervision_and_problem_formulation}
can benefit for network training and whether the proposed method can
be effective in using those samples.

\begin{figure}[H]
\begin{centering}
\includegraphics[width=12.4cm]{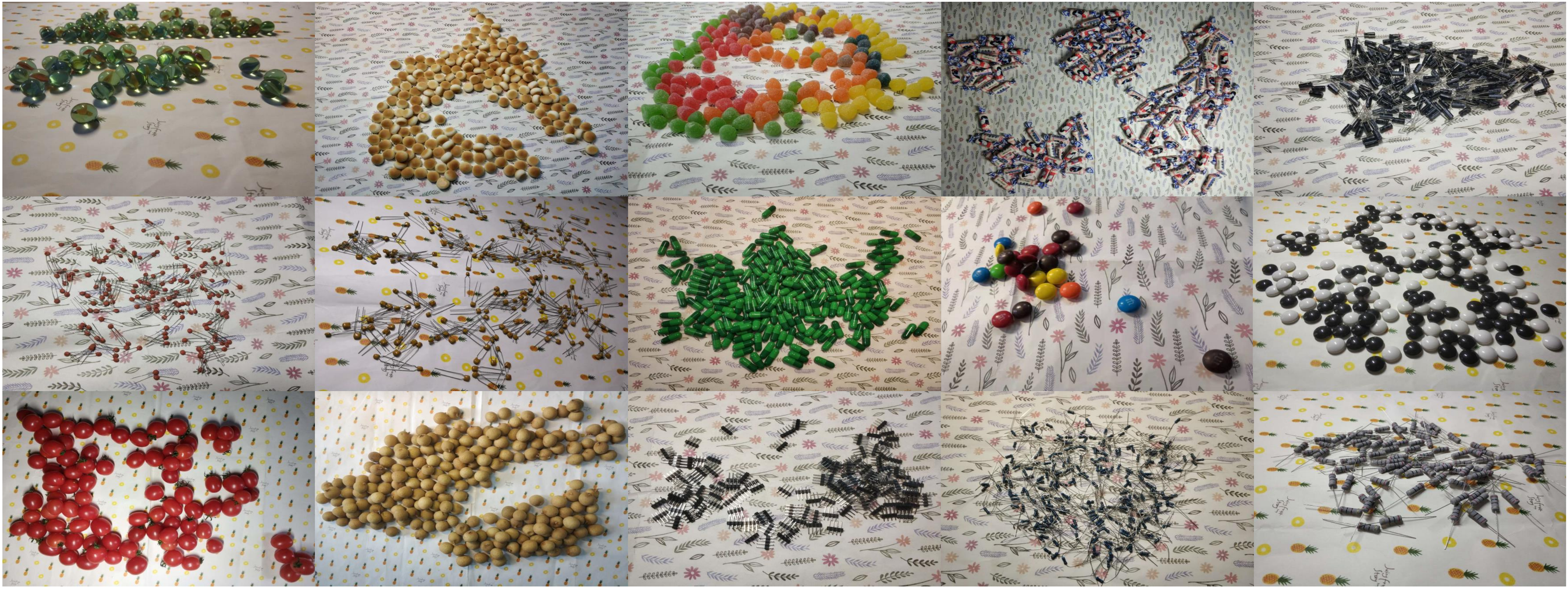}
\par\end{centering}
\caption{Typical images of the 15 categories from the MSCC dataset. \label{fig:Example-images-of-MSCC}}
\end{figure}

To address above issues, we collect a new dataset by following one
of the sample collection strategies discussed in Section \ref{sec:count_level_weak_supervision_and_problem_formulation}.
Specifically, we collect crowd object images for 15 different categories
(details see Figure \ref{fig:Example-images-of-MSCC}), including
Marble, Biscuit, two types of Candies (i.e., Candy1 and Candy2), three
types of Capacitors (i.e., Cap1, Cap2 and Cap3), Capsule, M\&M beans
(i.e., MM), Go, Cherry Tomato (i.e., Tomato), Longan, Pin-Header and
two types of Resistances (i.e., Res1 and Res2). For each category,
we choose a ``seed image'' and perform the location-level annotation
on it. This will give the total object count for the ``seed image''.

\begin{figure}[H]
\begin{centering}
\includegraphics[width=10.5cm]{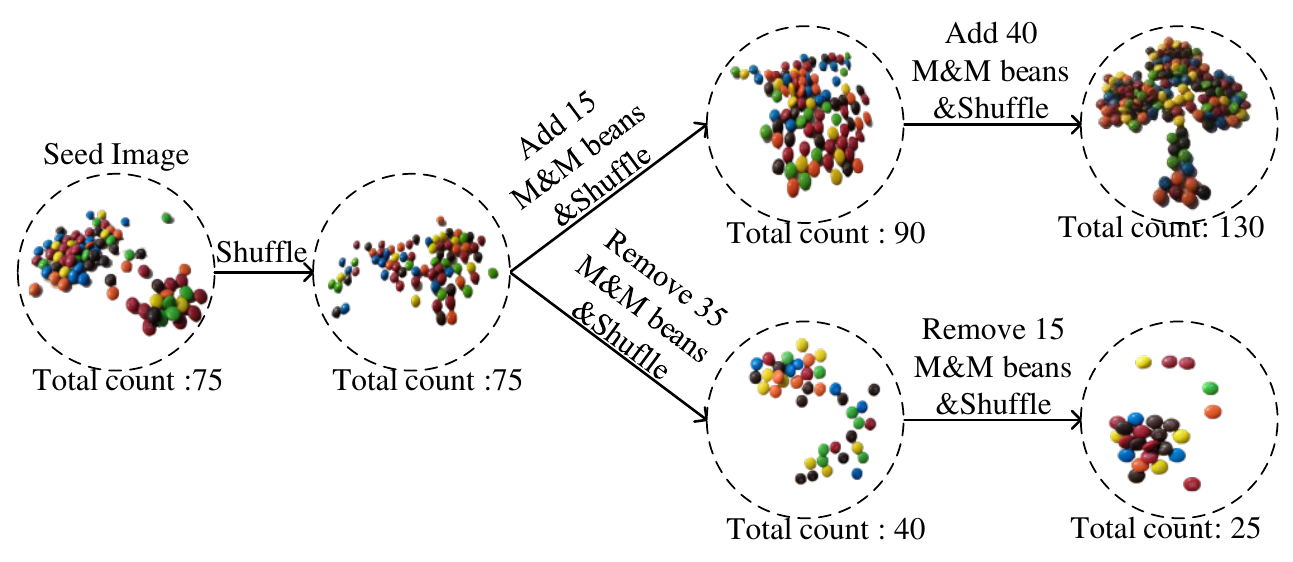}
\par\end{centering}
\caption{The procedure of generating the MSCC dataset. We obtain count-level
annotations from one ``seed image'', once we know the total count
of ``seed image''. When we shuffle the arrangement of the objects,
i.e., add or remove a small amount of objects before retaking, we
can infer the amount of objects of other images. \label{fig:Example-collect-images}}
\end{figure}

We collect more images by taking images from different viewpoints,
shuffling the object layout and changing the background before re-taking
the image. To collect various quantity levels in categories, we sequentially
remove or add a small amount of object. With those operations shown
in Figure \ref{fig:Example-collect-images}, we can quickly gather
a large number of images and infer their object count from the count
of the ``seed image''. In our dataset, we collect 200 images with
ten quantity levels per category. Since the images are collected by
taking multiple shots at the same pile of objects, we name our dataset
as Multi-Shot Crowd Counting (MSCC) dataset.

For each object category, we use 170 images as the training set, consisting
of 169 images with the count-level annotation (i.e., weakly-supervised
images) and one image with the location-level annotation (i.e., fully-supervised
image). The other 30 images with the count-level annotation are used
as the validation set. To build the test set, we manually annotated
100 additional images with a total count value. The total count value
in the test set is designed to be more diverse than that in the training
set. This enables us to test if our trained model can generalize to
images with a different object count.

\section{Experimental Results}

\subsection{Experimental Setting}

In this section, we demonstrate the effectiveness of the proposed
weakly-supervised crowd counting method by conducting experiments
on three traditional benchmark datasets: ShanghaiTech \cite{zhang2016single},
UCF\_CC\_50 \cite{Idrees_2013_CVPR} and WorldExpo'10 \cite{zhang2015cross}
datasets, along with the proposed MSCC dataset. Different from the
evaluation protocol of other location-level fully-supervised crowd
counting methods, we have modified their evaluation protocols to suit
for our count-level weakly-supervised setting. For each dataset, we
divide the original training images into a weakly-supervised part
(i.e., count-level annotation ) and fully-supervised part (i.e., location-level
annotation). Following the existing works \cite{zhang2015cross,li2018csrnet:},
we use the Mean Absolute Error (MAE) and Mean Square Error (MSE) as
the evaluation metrics for the above three traditional crowd counting
datasets.

Meanwhile, two measures are used as the evaluation metrics for the
proposed MSCC dataset, i.e., the MAE and the \textbf{Relative Error
Ratio (RER)} which is derived from MAE by dividing the total object
count. The RER is an indicator of relative error and it is meaningful
because we usually allow larger estimation error for a more crowded
scene and vice versa. For example, incorrectly predicting 10 objects
for an image with 1,000 objects is more favorable than incorrectly
predicting 10 objects for an image with only 20 objects. These three
metrics are defined as follows:
\begin{equation}
MAE=\frac{1}{\mathcal{N}}\sum_{i=1}^{\mathcal{N}}\left|Pre_{i}-GT_{i}\right|,\label{eq:mae}
\end{equation}
\begin{equation}
MSE=\sqrt{\frac{1}{\mathcal{N}}\sum_{i=1}^{\mathcal{N}}\left|Pre_{i}-GT_{i}\right|^{2}},\label{eq:mse}
\end{equation}
\begin{equation}
RER=\frac{1}{\mathcal{N}}\sum_{i=1}^{\mathcal{N}}\frac{|Pre_{i}-GT_{i}|}{GT_{i}},\label{eq:rer}
\end{equation}
where $\mathcal{N}$ is the number of test images, $Pre_{i}$ is the
predicted object count of image $I^{i}$ and $GT_{i}$ is the corresponding
ground-truth total count. The lower result of these metrics means
the better performance.

Three methods compared in our experiment are:
\begin{itemize}
\item \textbf{Baseline1} only utilizes the part of images with location-level
annotation to train the feature extractor $f_{b}$ and primary branch
$g_{0}$ directly. In our proposed settings, only a small set of images
are provided with location-level annotation, it can be seen as a ``reduced''
training set sampled from the original training set. 
\item \textbf{Baseline2} the straightforward solution to incorporate the
count-level weak supervision in Subsection \ref{sect:naive_solution_for_weakly_supervised_solution}.
In our network, this method is equivalent to only training a single
branch: the feature extractor $f_{b}$ and the primary branch $g_{0}$.
Different from the ``Baseline1'' method, it jointly considers a
small mount of location-level annotations and a large mount of count-level
annotations. Straightly we require training image $I^{m}$ with location-level
annotation generate density map close to ground-truth density map
$D_{m}$. For training image $I^{n}$ with count-level annotation,
we require the integral of predicted density map close to ground-truth
total count $c_{n}$.
\item \textbf{MATT} is the proposed method that leverages multiple auxiliary
tasks for model training. Notice that \textquotedblleft Baseline1\textquotedblright{}
and \textquotedblleft Baseline2\textquotedblright{} only leverage
single branch: the feature extractor $f_{b}$ and primary branch $g_{0}$.
For the sake of train a more robust $f_{b}$, we propose the MATT
method. When the model meets image $I^{j}$ with count-level weak
supervision, firstly we introduce four auxiliary branches to predict
density map, then according to Algorithm \ref{algo:training} we update
the corresponding parameters. The auxiliary branch can promote the
training of $f_{b}$.
\end{itemize}
The differences of the above three methods are demonstrated in Table
\ref{tab:comparison-three-method}. Note that, during testing we discard
$g_{1}$,...,$g_{k}$ and only use $f_{b}$ and $g_{0}$ for the final
density maps.

\begin{table}[h!]
\centering{}\caption{The comparison between three methods. \label{tab:comparison-three-method}}
\begin{tabular}{cccc}
\hline 
Train method & Annotated type & Loss function & Parameters in model\tabularnewline
\hline 
\multirow{2}{*}{Baseline1} & location-level & MSE Loss & $f_{b}$ and $g_{0}$\tabularnewline
\cline{2-4} \cline{3-4} \cline{4-4} 
 & count-level & - & -\tabularnewline
\hline 
\multirow{2}{*}{Baseline2} & location-level & MSE Loss & $f_{b}$ and $g_{0}$\tabularnewline
\cline{2-4} \cline{3-4} \cline{4-4} 
 & count-level & Count loss & $f_{b}$ and $g_{0}$\tabularnewline
\hline 
\multirow{3}{*}{MATT} & location-level & MSE loss & $f_{b}$ and $g_{0}$\tabularnewline
\cline{2-4} \cline{3-4} \cline{4-4} 
 & \multirow{2}{*}{count-level} & Count loss & $f_{b}$ and $g_{0}$\tabularnewline
\cline{3-4} \cline{4-4} 
 &  & Auxiliary loss & $f_{b}$ and $g_{1}$,...,$g_{k}$\tabularnewline
\hline 
\end{tabular}
\end{table}

\subsection{Implementation Details\label{sec:implementation_details}}

Our feature extractor is realized based on the CSRNet \cite{li2018csrnet:},
which is one of the state-of-the-art solutions for crowd counting.
Certainly, the proposed method can be extended to other feature extractor
architectures. Specifically, the layers to the third-last layer of
CSRNet are used as the backbone which produces a feature map with
256 channels. One primary branch and four auxiliary branches are used
in our implementation. Each branch contains three convolutional layers,
the kernel size of each convolutional layer is $3\times3$ and the
channel number reduced from 256 to 128, 64 and finally 1. For four
auxiliary branches, different convolutional kernels $h_{k}$ are used
to construct the loss in Eq. \ref{eq:auxiliary-loss}. We choose them
as predefined Gaussian kernels with different covariance matrices
(see details in Subsection \ref{sec:multiple_auxiliary_task_training})\textcolor{blue}{.}
The PyTorch \cite{paszke2017automatic} is applied to implement the
proposed method. Adam \cite{Kingma_2014} is used as the optimizer.
All the hyper-parameters are chosen by using the validation set. We
will release the source code and the MSCC dataset upon the acceptance
of this work.

\subsection{Datasets and Results}

In this subsection, comprehensive experiments are conducted\textcolor{red}{{}
}to evaluate the performance of the proposed method. Please note that,
our method only uses a few training images with location-level annotation
and a large amount of training images with count-level annotation,
which is only comparable to methods in the ``weakly'' part. However
we also list the fully-supervised crowd counting state-of-the-art
results in the ``fully'' part just for reference in Table \ref{tab:The-performance-ShanghaiTech}-\ref{tab:The-performance-comparison-WorldExpo}.

\subsubsection{Evaluation on the ShanghaiTech dataset}

The ShanghaiTech dataset is a large-scale crowd counting dataset containing
1,198 images with 330,165 annotated heads \cite{zhang2016single},
which is divided into two parts. Part\_A contains 482 images with
241,667 annotated heads, and 300 images are used for training while
the rest for testing. In our experiment, we only use the count-level
annotation for 270 images from the training set and the object location-level
annotation from the remaining 30 images. Part\_B has 716 images with
88,498 annotated heads taken from street scenes in the Shanghai city,
with 400 images for training. Similar to Part\_A, we only use count-level
weak annotation for 380 images from the training set and use location-level
annotation for the remaining 20 images. This modification suits for
the evaluation in the weakly-supervised setting.

\begin{figure}
\begin{centering}
\includegraphics[width=12.2cm]{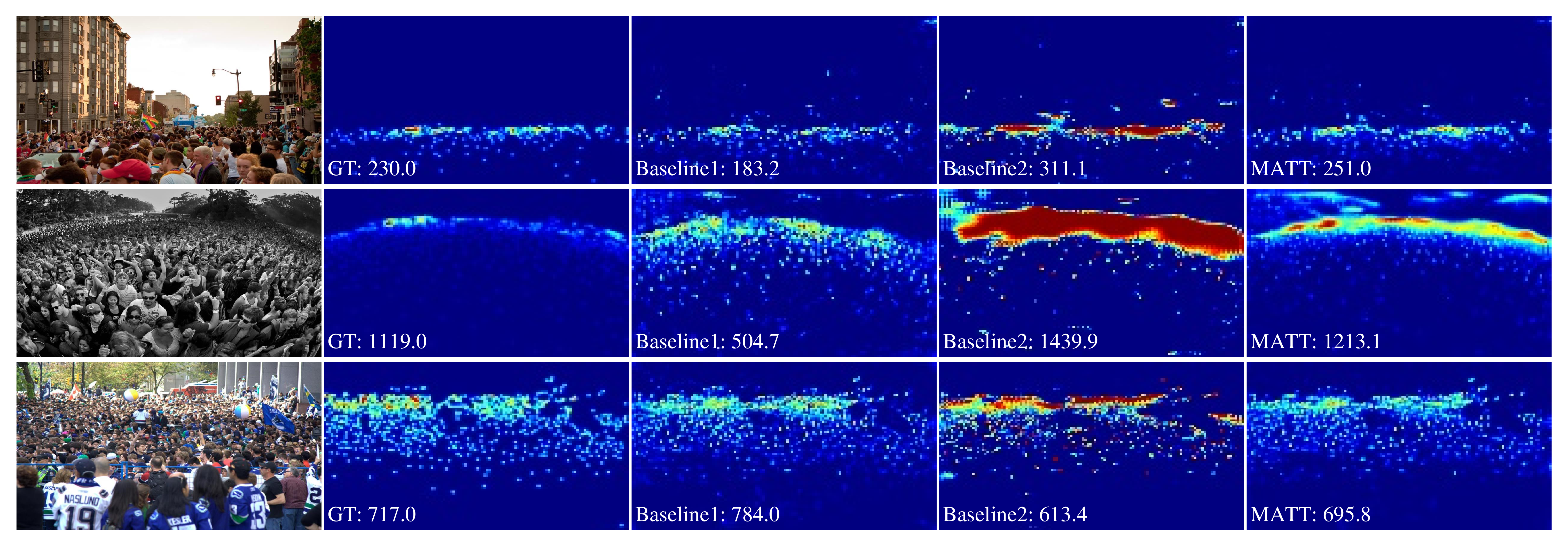}
\par\end{centering}
\caption{A comparison of the density map generated by different methods of
the proposed method on the ShanghaiTech Part\_A dataset. \label{fig:A-comparison-of-density-map-generation}}
\end{figure}

The experimental results are shown in Table \ref{tab:The-performance-ShanghaiTech}.
In the traditional fully-supervised setting, works based on CSRNet
\cite{li2018csrnet:} leverage attention mechanism \cite{liu2019adcrowdnet,valloli2019wnet}
and context information \cite{liu2019context} can regress more accurate
density maps and obtain lower MAE in the test stage. While our method
is comparable to those methods. In addition, in the weak supervision
setting, if we train our network only with the location-level annotations
(i.e. ``Baseline1''). The performance is not satisfying, and it
only achieves a MAE of 106.5 on Part\_A and a MAE of 16.4 on Part\_B.
From the results reported in CSRNet \cite{li2018csrnet:}, if the
location-level annotations are used for all the samples, the performance
of the same network will achieve a MAE of 68.2 and 10.6 on Part\_A
and Part\_B respectively. Thus, reducing the number of location-level
annotations incurs a significant performance drop due to a lack of
location-level annotated images. By using the naive method of incorporating
count-level supervision, the performance is significantly improved.
In Part\_A, the MAE is reduced to 89.3. The proposed method can also
make a significant improvement over the ``Baseline1'', and the improvement
is much larger than the naive method. Comparing with ``Baseline2'',
we achieve an improvement by 9.2 heads in the terms of MAE on Part\_A.
On Part\_B, our method also reduces the MAE by around 1.8 heads. The
MAE improvement is smaller because Part\_B contains fewer crowd scenes,
and the MAE differences between various methods tend to be smaller.
It is easy to calculate that each image in Part\_A contains 501.4
persons averagely as four times as Part\_B nearly. In other words,
Part\_A is more challenging than Part\_B. This is also a trend observed
in the works in the location-level supervised setting mentioned in
Table \ref{tab:The-performance-ShanghaiTech}. In Figure \ref{fig:A-comparison-of-density-map-generation},
we also visualize the estimated density maps of different methods.
It is interesting to find that the density map generated by the proposed
method is more similar to the ground-truth density map. In comparison,
the density map generated by the naive method tends to exhibit large
homogeneous regions.

\begin{table}[h!]
\centering{}\caption{The comparison on the ShanghaiTech dataset. The best results are in
underline and bold font for ``Fully'' and ``Weakly'' methods respectively.
\label{tab:The-performance-ShanghaiTech}}
\begin{tabular}{cccccc}
\hline 
\multirow{1}{*}{} & \multirow{1}{*}{Method} & \multicolumn{2}{c}{Part\_A} & \multicolumn{2}{c}{Part\_B}\tabularnewline
\hline 
 &  & MAE & MSE & MAE & MSE\tabularnewline
\hline 
\multirow{10}{*}{Fully} & MCNN \cite{zhang2016single} & 110.2 & 173.2 & 26.4 & 41.3\tabularnewline
\cline{2-6} \cline{3-6} \cline{4-6} \cline{5-6} \cline{6-6} 
 & Switch-CNN \cite{Sam_2017} & 90.4 & 135.0 & 21.6 & 33.4\tabularnewline
\cline{2-6} \cline{3-6} \cline{4-6} \cline{5-6} \cline{6-6} 
 & ACSCP \cite{shen2018crowd} & 75.7 & 102.7 & 17.2 & 27.4\tabularnewline
\cline{2-6} \cline{3-6} \cline{4-6} \cline{5-6} \cline{6-6} 
 & CP-CNN \cite{Sindagi_2017} & 73.6 & 106.4 & 20.1 & 30.1\tabularnewline
\cline{2-6} \cline{3-6} \cline{4-6} \cline{5-6} \cline{6-6} 
 & CSRNet \cite{li2018csrnet:} & 68.2 & 115.0 & 10.6 & 16.0\tabularnewline
\cline{2-6} \cline{3-6} \cline{4-6} \cline{5-6} \cline{6-6} 
 & SANet \cite{Cao_2018_ECCV} & 67.0 & 104.5 & 8.4 & 13.6\tabularnewline
\cline{2-6} \cline{3-6} \cline{4-6} \cline{5-6} \cline{6-6} 
 & ADCrowdNet \cite{liu2019adcrowdnet} & 63.2 & 98.9 & 8.2 & 15.7\tabularnewline
\cline{2-6} \cline{3-6} \cline{4-6} \cline{5-6} \cline{6-6} 
 & CAN \cite{liu2019context} & 62.3 & 100 & 7.8 & 12.2\tabularnewline
\cline{2-6} \cline{3-6} \cline{4-6} \cline{5-6} \cline{6-6} 
 & SFANet \cite{zhu2019dual} & 59.8 & 99.3 & \uline{6.9} & 10.9\tabularnewline
\cline{2-6} \cline{3-6} \cline{4-6} \cline{5-6} \cline{6-6} 
 & W-NET \cite{valloli2019wnet} & \uline{59.5} & \uline{97.3} & \uline{6.9} & \uline{10.3}\tabularnewline
\hline 
\multirow{3}{*}{Weakly} & Baseline1 & 106.5 & 167.4 & 16.4 & 25.7\tabularnewline
\cline{2-6} \cline{3-6} \cline{4-6} \cline{5-6} \cline{6-6} 
 & Baseline2 & 89.3 & 135.8 & 13.5 & 19.9\tabularnewline
\cline{2-6} \cline{3-6} \cline{4-6} \cline{5-6} \cline{6-6} 
 & MATT & \textbf{80.1} & \textbf{129.4} & \textbf{11.7} & \textbf{17.5}\tabularnewline
\hline 
\end{tabular}
\end{table}

\subsubsection{Evaluation on the UCF\_CC\_50 Dataset}

The UCF\_CC\_50 dataset \cite{Idrees_2013_CVPR} contains only 50
black and white images which are considered to be challenging due
to the high object density in the images. Its count value varies from
94 to 4,543. In our weakly-supervised setting, 5 images are with location-level
annotation and the rest 45 images are with only count-level annotation.
20 patches with half image size are cropped from each image, and 5-fold
cross-validation is used to evaluate the proposed method \cite{Idrees_2013_CVPR}.
4 location-level annotated images with ground-truth location-level
supervision and 36 count-level annotated images with ground-truth
count-level supervision are randomly selected to form the training
set.

The performance of the compared methods is shown in Table \ref{tab:The-performance-on-UCFF}.
As seen, with a small amount of location-level supervised samples,
the crowd counting performance is poor. With the count-level supervision,
even the naive method can reduce the MAE from 461.4 to 405. Our proposed
method again shows a significant improvement over the naive method.
It improves the MAE almost by 50 heads comparing with the naive method.
This clearly demonstrates the advantage of the proposed method.

\begin{table}[h!]
\centering{}\caption{The performance comparison on the UCFF\_CC\_50 dataset. The best results
are in underline and bold font for ``Fully'' and ``Weakly'' methods
respectively. \label{tab:The-performance-on-UCFF}}
\begin{tabular}{cccc}
\hline 
\multirow{1}{*}{} & Method & MAE & MSE\tabularnewline
\hline 
\multirow{10}{*}{Fully} & MCNN \cite{zhang2016single} & 377.6 & 509.1\tabularnewline
\cline{2-4} \cline{3-4} \cline{4-4} 
 & Switch-CNN \cite{Sam_2017} & 318.1 & 439.2\tabularnewline
\cline{2-4} \cline{3-4} \cline{4-4} 
 & CP-CNN \cite{Sindagi_2017} & 295.8 & 320.9\tabularnewline
\cline{2-4} \cline{3-4} \cline{4-4} 
 & ACSCP \cite{shen2018crowd} & 291.0 & 404.6\tabularnewline
\cline{2-4} \cline{3-4} \cline{4-4} 
 & CSRNet \cite{li2018csrnet:} & 266.1 & 397.5\tabularnewline
\cline{2-4} \cline{3-4} \cline{4-4} 
 & ADCrowdNet \cite{liu2019adcrowdnet} & 257.1 & 363.5\tabularnewline
\cline{2-4} \cline{3-4} \cline{4-4} 
 & SANet \cite{Cao_2018_ECCV} & 258.4 & 334.9\tabularnewline
\cline{2-4} \cline{3-4} \cline{4-4} 
 & SFANet \cite{zhu2019dual} & 219.6 & 316.2\tabularnewline
\cline{2-4} \cline{3-4} \cline{4-4} 
 & CAN \cite{liu2019context} & 212.2 & \uline{243.7}\tabularnewline
\cline{2-4} \cline{3-4} \cline{4-4} 
 & W-Net\cite{valloli2019wnet} & \uline{201.9} & 309.2\tabularnewline
\hline 
\multirow{3}{*}{Weakly} & Baseline1 & 461.4 & 779.9\tabularnewline
\cline{2-4} \cline{3-4} \cline{4-4} 
 & Baseline2 & 405.0 & 586.2\tabularnewline
\cline{2-4} \cline{3-4} \cline{4-4} 
 & MATT & \textbf{355.0} & \textbf{550.2}\tabularnewline
\hline 
\end{tabular}
\end{table}

\subsubsection{Evaluation on the WorldExpo'10 Dataset}

This dataset contains 3,980 uniformly sampled frames from video sequences
captured by 108 surveillance cameras from Shanghai 2010 WorldExpo
\cite{zhang2015cross}. 3,380 frames are used for training while
the remaining 600 frames for testing. In our setting, 5\% frames (169),
in the training set are used as the location-level annotated images
while the remaining 95\% frames (3,211) are used as count-level annotated
images.

The results are shown in Table \ref{tab:The-performance-comparison-WorldExpo}.\textbf{
}From the results, We can see that our proposed method is superior
to the baselines. The proposed method achieve the highest average
MAE, improving more than 2.4 heads over the ``Baseline1''. For the
performance in each section, our method attains 4 lowest MAE among
all 5 scenarios. In comparison, the ``Baseline2'' does not improve
too much over the ``Baseline1''. This indicates an appropriate learning
method can play an important role when using the count-level annotations.

\begin{table}[th!]
\centering{}\caption{The performance comparison in terms of MAE on the WorldExpo'10 dataset.
It can be seen that the performance of the proposed method MATT is
close to the performance of methods that all training images using
location-level supervision. The best results are in underline and
bold font for ``Fully'' and ``Weakly'' methods respectively. \label{tab:The-performance-comparison-WorldExpo}}
{\scriptsize{} }%
\begin{tabular}{cccccccc}
\hline 
\multirow{1}{*}{} & Method & Sce.1 & Sce.2 & Sce.3 & Sce.4 & Sce.5 & Avg.\tabularnewline
\hline 
\multirow{8}{*}{Fully} & MCNN \cite{zhang2016single} & 3.4 & 20.6 & 12.9 & 13.0 & 8.1 & 11.6\tabularnewline
\cline{2-8} \cline{3-8} \cline{4-8} \cline{5-8} \cline{6-8} \cline{7-8} \cline{8-8} 
 & Switch-CNN \cite{Sam_2017} & 4.4 & 15.7 & 10.0 & 11.0 & 5.9 & 9.4\tabularnewline
\cline{2-8} \cline{3-8} \cline{4-8} \cline{5-8} \cline{6-8} \cline{7-8} \cline{8-8} 
 & CP-CNN \cite{Sindagi_2017} & 2.9 & 14.7 & 10.5 & 10.4 & 5.8 & 8.9\tabularnewline
\cline{2-8} \cline{3-8} \cline{4-8} \cline{5-8} \cline{6-8} \cline{7-8} \cline{8-8} 
 & CSRNet \cite{li2018csrnet:} & 2.9 & 11.5 & \uline{8.6} & 16.6 & 3.4 & 8.6\tabularnewline
\cline{2-8} \cline{3-8} \cline{4-8} \cline{5-8} \cline{6-8} \cline{7-8} \cline{8-8} 
 & SANet \cite{Cao_2018_ECCV} & 2.6 & 13.2 & 9.0 & 13.3 & 3.0 & 8.2\tabularnewline
\cline{2-8} \cline{3-8} \cline{4-8} \cline{5-8} \cline{6-8} \cline{7-8} \cline{8-8} 
 & ACSCP \cite{shen2018crowd} & 2.8 & 14.1 & 9.6 & \uline{8.1} & 2.9 & 7.5\tabularnewline
\cline{2-8} \cline{3-8} \cline{4-8} \cline{5-8} \cline{6-8} \cline{7-8} \cline{8-8} 
 & ADCrowdNet \cite{liu2019adcrowdnet} & \uline{1.6} & 13.2 & 8.7 & 10.6 & \uline{2.6} & 7.3\tabularnewline
\cline{2-8} \cline{3-8} \cline{4-8} \cline{5-8} \cline{6-8} \cline{7-8} \cline{8-8} 
 & CAN \cite{liu2019context} & 2.4 & \uline{9.4} & 8.8 & 11.2 & 4.0 & \uline{7.2}\tabularnewline
\hline 
\multirow{3}{*}{Weakly} & Baseline1 & 4.2 & 18.3 & 11.9 & 22.8 & \textbf{3.3} & 12.1\tabularnewline
\cline{2-8} \cline{3-8} \cline{4-8} \cline{5-8} \cline{6-8} \cline{7-8} \cline{8-8} 
 & Baseline2 & 3.9 & 17.6 & 16.1 & 16.3 & 4.8 & 11.7\tabularnewline
\cline{2-8} \cline{3-8} \cline{4-8} \cline{5-8} \cline{6-8} \cline{7-8} \cline{8-8} 
 & MATT & \textbf{3.8} & \textbf{13.1} & \textbf{10.4} & \textbf{15.9} & 5.3 & \textbf{9.7}\tabularnewline
\hline 
\end{tabular}
\end{table}

\subsubsection{Evaluation on the MSCC Dataset}

Finally, we compare our method on the proposed MSCC dataset. We strictly
follow the experimental protocol in Section \ref{sec:multi_shot_crowd_counting}.
The results are shown in Table \ref{tab:The-performance-comparison-MSCC}.
As seen, for all categories, the proposed method achieves the best
MAE and RER, especially for categories with irregularly shaped objects
such as pin headers. For this category, the proposed method reduces
MAE from 27.9 to 5.8 and delivers 87.8\% lower RER. On average, the
proposed method achieves a MAE of 6.8 and a RER of 15.7. It is 63.8\%
lower MAE and 76.4\% lower RER than a MAE of 18.8 and a RER of 66.5
from the ``Baseline1'' method.

Note that the object count annotation on the MSCC dataset is collected
with less effort. The better performance of the ``Baseline2'' and
``MATT'' comparing with the ``Baseline1'' clearly demonstrates
that using the effortlessly collected count-level weakly supervised
images is beneficial for training a crowd counting network. Also,
the consistently superior performance of the proposed approach over
the naive method demonstrates that our method is useful in practice.

\begin{table}
\centering{}\caption{The performance comparison for different methods on the MSCC dataset.
The best results are in bold font. \label{tab:The-performance-comparison-MSCC}}
\begin{tabular}{cccccccc}
\cline{1-7} \cline{2-7} \cline{3-7} \cline{4-7} \cline{5-7} \cline{6-7} \cline{7-7} 
\multicolumn{2}{c}{} & Marble & Biscuit & Candy1 & Candy2 & Cap1 & \multirow{1}{*}{}\tabularnewline
\cline{1-7} \cline{2-7} \cline{3-7} \cline{4-7} \cline{5-7} \cline{6-7} \cline{7-7} 
Baseline1 & \multirow{3}{*}{MAE\textbackslash RER} & 8.7\textbackslash 26.3 & 11.4\textbackslash 41.4 & 16.2\textbackslash 86.7 & 26.7\textbackslash 52.3 & 34.5\textbackslash 180.8 & \tabularnewline
\cline{1-1} \cline{3-7} \cline{4-7} \cline{5-7} \cline{6-7} \cline{7-7} 
Baseline2 &  & 6.2\textbackslash 22.2 & 6.0\textbackslash 12.3 & 12.0\textbackslash 28.7 & 8.7\textbackslash 17.0 & 9.3\textbackslash 28.1 & \tabularnewline
\cline{1-1} \cline{3-7} \cline{4-7} \cline{5-7} \cline{6-7} \cline{7-7} 
MATT &  & \textbf{4.8\textbackslash 14.3} & \textbf{3.9}\textbackslash\textbf{9.4} & \textbf{8.4}\textbackslash\textbf{19.7} & \textbf{5.4\textbackslash 11.7} & \textbf{7.7}\textbackslash\textbf{23.3} & \tabularnewline
\cline{1-7} \cline{2-7} \cline{3-7} \cline{4-7} \cline{5-7} \cline{6-7} \cline{7-7} 
\multicolumn{2}{c}{} & Cap2 & Cap3 & Capsule & MM & Go & \tabularnewline
\cline{1-7} \cline{2-7} \cline{3-7} \cline{4-7} \cline{5-7} \cline{6-7} \cline{7-7} 
Baseline1 & \multirow{3}{*}{MAE\textbackslash RER} & 21.6\textbackslash 86.5 & 23.9\textbackslash 86.7 & 19.9\textbackslash 119.1 & 8.1\textbackslash 30.9 & 9.6\textbackslash 23.3 & \tabularnewline
\cline{1-1} \cline{3-7} \cline{4-7} \cline{5-7} \cline{6-7} \cline{7-7} 
Baseline2 &  & 7.7\textbackslash 16.9 & 10.6\textbackslash 22.7 & 11.1\textbackslash 20.4 & 14.9\textbackslash 33.2 & 8.8\textbackslash 18.7 & \tabularnewline
\cline{1-1} \cline{3-7} \cline{4-7} \cline{5-7} \cline{6-7} \cline{7-7} 
MATT &  & \textbf{6.4}\textbackslash\textbf{14.3} & \textbf{7.2}\textbackslash\textbf{17.4} & \textbf{4.3}\textbackslash\textbf{9.8} & \textbf{5.4}\textbackslash\textbf{12.6} & \textbf{8.3\textbackslash 18.4} & \tabularnewline
\hline 
 &  & Tomato & Longan & Pin-Header & Res1 & Res2 & \multicolumn{1}{c}{Avg.}\tabularnewline
\hline 
Baseline1 & \multirow{3}{*}{MAE\textbackslash RER} & 13.8\textbackslash 39.7 & 5.4\textbackslash 16.0 & 27.9\textbackslash 103.8 & 20.0\textbackslash 45.4 & 34.5\textbackslash 59.8 & \multicolumn{1}{c}{18.8\textbackslash 66.5}\tabularnewline
\cline{1-1} \cline{3-8} \cline{4-8} \cline{5-8} \cline{6-8} \cline{7-8} \cline{8-8} 
Baseline2 &  & 11.3\textbackslash 35.2 & 10.8\textbackslash 27.2 & 15.1\textbackslash 26.7 & 23.0\textbackslash 35.9 & 15.0\textbackslash 31.6 & \multicolumn{1}{c}{11.4\textbackslash 25.1}\tabularnewline
\cline{1-1} \cline{3-8} \cline{4-8} \cline{5-8} \cline{6-8} \cline{7-8} \cline{8-8} 
MATT &  & \textbf{7.3\textbackslash 21.9} & \textbf{4.0}\textbackslash\textbf{10.6} & \textbf{5.8}\textbackslash\textbf{12.7} & \textbf{15.8}\textbackslash\textbf{24.2} & \textbf{7.2}\textbackslash\textbf{14.6} & \multicolumn{1}{c}{\textbf{6.8}\textbackslash\textbf{15.7}}\tabularnewline
\hline 
\end{tabular}
\end{table}

\subsection{Ablation Study}

\subsubsection{Other Alternative Auxiliary Loss Functions}

The auxiliary loss in this paper involves two terms for count-level
annotated images: a count loss term and a density map consistency
loss. To examine their impacts, we conduct an ablation study on the
ShanghaiTech Part\_A dataset by constructing two alternative variations
of the proposed method. The first variation only uses a count loss.
The second variation does not use the total count loss but only uses
the first term in Eq. \ref{eq:auxiliary-loss}. The asymmetry training
strategy is also used in those variations.

\begin{table}[h!]
\centering{}\caption{Results of alternative constructions of the auxiliary loss. Evaluated
on the ShanghaiTech Part\_A dataset. The best results are in bold
font. \label{tab:Results-of-Auxiliary-loss}}
\begin{tabular}{ccc}
\hline 
Method & MAE & MSE\tabularnewline
\hline 
only count loss & 85.7 & 138.0\tabularnewline
\hline 
only MSE loss & 83.3 & 132.6\tabularnewline
\hline 
MATT & \textbf{80.1} & \textbf{129.4}\tabularnewline
\hline 
\end{tabular}
\end{table}

The performance comparison is shown in Table \ref{tab:Results-of-Auxiliary-loss}.
As seen, the proposed method achieves better performance than the
first variation which can be seen as a multi-branch variation of ``Baseline2''.
In comparison, the second variation achieves better performance and
is even comparable to the performance of MATT. This suggests that
MSE loss is the most effective part of the total loss function in
the proposed method. Combining both loss terms can achieve the best
performance.

\subsubsection{Varying the Number of Auxiliary Branches}

\begin{figure}
\begin{centering}
\includegraphics[width=8.5cm]{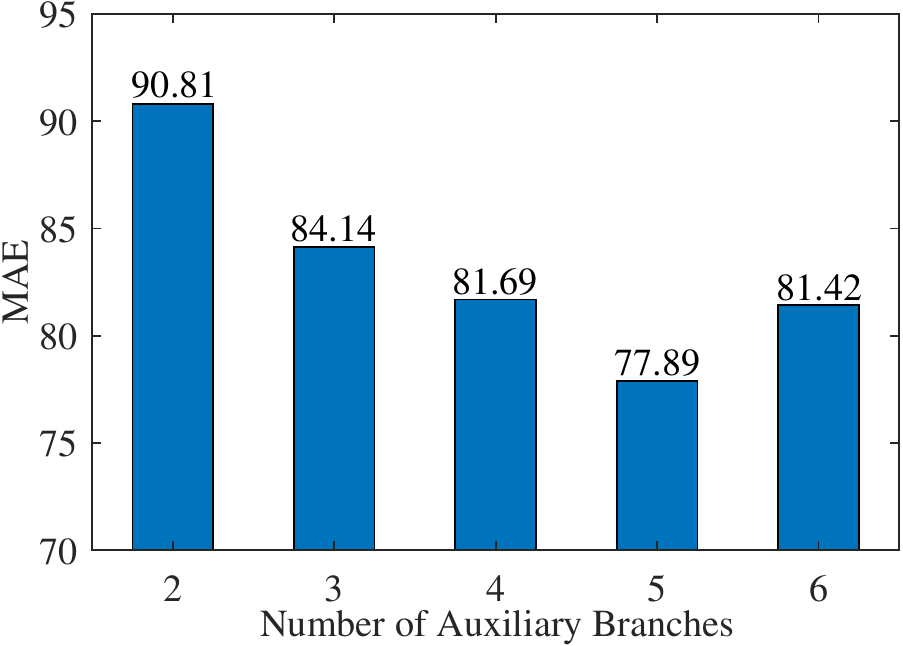}
\par\end{centering}
\caption{Impact of different number of auxiliary branches. Evaluated in terms
of MAE on the ShanghaiTech Part\_A dataset. \label{fig:Results-of-different-number-of-branches}}
\end{figure}

In the proposed method, we use four auxiliary branches for the simplicity
of implementation. It would be interesting to see if using more or
fewer branches will result in better or poor performance. The experimental
results on Shanghai Tech Part\_A dataset are presented in Figure \ref{fig:Results-of-different-number-of-branches}.
It can be observed that if only two auxiliary branches are used, the
performance will not be better than the naive method --- both methods
achieve MAE around 90.8 heads. However, with the increase in the number
of branches, the estimation error is reduced. The experiment results
show that the best performance is achieved when using 4 to 6 auxiliary
branches. From those results, it is clear to see the importance of
using more branches to maintain the diversity of the auxiliary supervision
signals.

\subsubsection{The Importance of the Asymmetry Training}

Our work uses an asymmetry training strategy to learn the auxiliary
tasks, and the auxiliary tasks only learn from the primary task while
the primary task will not learn from the auxiliary tasks. In our strategy,
the primary task is used for all images and learn from the discrepancy
between predicted density map $F_{0}(\mathbf{x})$ and ground-truth
density map $D_{m}(\mathbf{x})$ or ground-truth total count $c_{n}$.

The auxiliary task works only when taking count-level annotated images
$I_{n}$. Learning from the discrepancy between predicted density
map $F_{k}(\mathbf{x})$ and (1) corresponding density map $F_{0}(\mathbf{x})$
generated at the primary branch, (2) ground-truth total count $c_{n}$.
We postulate that this is important because otherwise, the auxiliary
supervisions signals will flow into the primary branch. Because the
auxiliary branch will fit a less accurate density map --- the predicted
density map from the primary task, the supervisions signals from the
auxiliary task is inevitably noisy. Thus, it might be harmful to allow
its gradient to update the primary branch. In this study, we conduct
an experiment to verify our postulation on the ShanghaiTech Part\_A
dataset. As shown in Table \ref{tab:Results-of-different-asymmetry},
after removing the asymmetry training strategy, we only obtained the
MAE of 89.0 and MSE of 140.6 on the ShanghaiTech Part\_A dataset.
This is much worse than its original version and this validates our
asymmetry training postulation.

\begin{table}[h!]
\centering{}\caption{Impact of the asymmetry training strategy. Evaluated on the ShanghaiTech
Part\_A dataset. The best results are in bold font. \label{tab:Results-of-different-asymmetry}}
\begin{tabular}{ccc}
\hline 
Method & MAE & MSE\tabularnewline
\hline 
symmetry & 89.0 & 140.6\tabularnewline
\hline 
asymmetry & \textbf{80.1} & \textbf{129.4}\tabularnewline
\hline 
\end{tabular}
\end{table}

\subsubsection{MATT for Fully Annotated Data}

The proposed method is in a spirit similar to the multi-task learning
and one may wonder the good performance of our method essentially
comes from multi-task learning. In other words, our method may lead
to better baseline performance even with the fully-annotated part
of training data. To verify this, we apply the proposed MATT method
on the fully annotated part of the training set on the ShanghaiTech
Part\_A dataset, and report the results in Table \ref{tab:Results-of-auxiliary-branches}.
Interestingly, using MATT for a location-level annotated image only
performs slight better than ``Baseline1'', which only includes the
primary branch (i.e., single task learning). This indicates that the
superior performance of the proposed method is largely originated
from its ability to better exploiting the information provided in
the count-level annotated image not only the multi-task learning.

\begin{table}[h!]
\centering{}\caption{Results of using proposed method for training based on the fully annotated
part of data on the ShanghaiTech Part\_A dataset. The best results
are in bold font. \label{tab:Results-of-auxiliary-branches}}
\begin{tabular}{ccc}
\hline 
Method & MAE & MSE\tabularnewline
\hline 
MATT on Fully & \textbf{104.6} & \textbf{162.7}\tabularnewline
\hline 
Baseline1 & 106.5 & 167.4\tabularnewline
\hline 
\end{tabular}
\end{table}

\section{Conclusions}

In this paper, we point out that the count-level annotations can be
easily obtained. We also identify the limitation of the straightforward
method to weakly supervised counting and propose a novel Multiple
Auxiliary Task Training (MATT) scheme to learn a better crowd counting
model. To verify the benefits of count-level weakly supervised learning,
we also introduce a new dataset. By performing experiments on the
newly introduced dataset and the traditional publicly available crowd
counting datasets, we demonstrate that the proposed method is superior
to the straightforward weakly supervised crowd counting method and
can leverage the count-level weak supervision to build a better crowd
counting model.

\section*{Acknowledgment}

This work was supported by the Key Research and Development Program
of Sichuan Province (2019YFG0409). Lingqiao Liu was in part supported
by ARC DECRA Fellowship DE170101259.

\bibliographystyle{elsarticle-num}
\bibliography{reference}

\end{document}